%% file: root.tex
\documentclass[letterpaper, 10 pt, conference]{ieeeconf}  

\IEEEoverridecommandlockouts                              

\usepackage[dvipsnames]{xcolor}
\usepackage[pdftex]{graphicx} 
\usepackage{epsfig} 

\usepackage{amsmath} 
\usepackage{float}
\usepackage[notransparent]{svg}
\usepackage{todonotes}
\usepackage{amssymb}
\usepackage{afterpage}
\usepackage{url}
\usepackage{fancyhdr}
\usepackage[headheight=23pt]{geometry}

\newcommand{\PREPRINTYEAR}{2025}
\newcommand{\PUBLISHEDIN}{IEEE International Conference on Robotics and Automation}

\usepackage[placement=top,vshift=-7]{background}
\SetBgScale{0.8}
\SetBgContents{\copyright{} \PREPRINTYEAR~\PUBLISHEDIN. PREPRINT VERSION - DO NOT DISTRIBUTE.}
\SetBgColor{black}
\SetBgAngle{0}
\SetBgOpacity{1.0}

\usepackage{tikz}

\newcommand\copyrighttext{%
  \footnotesize \textcopyright 2025 IEEE.  Personal use of this material is permitted.  Permission from IEEE must be obtained for all other uses, in any current or future media, including reprinting/republishing this material for advertising or promotional purposes, creating new collective works, for resale or redistribution to servers or lists, or reuse of any copyrighted component of this work in other works.}
\newcommand\copyrightnotice{%
\begin{tikzpicture}[remember picture,overlay]
\node[anchor=south,yshift=10pt] at (current page.south) {\fbox{\parbox{\dimexpr\textwidth-\fboxsep-\fboxrule\relax}{\copyrighttext}}};
\end{tikzpicture}%
}

\title{\LARGE \bf
Manual, Semi or Fully Autonomous Flipper Control? A Framework for Fair Comparison}

\author{Valent\'yn {\v C}\'ihala, Martin Pecka, Tom\'a{\v s} Svoboda and Karel Zimmermann
\thanks{All authors are with Czech Technical University in Prague, Faculty of Electrical Engineering, Department of Cybernetics. Corresponding author V. {\v C}\'ihala:  cihalval@fel.cvut.cz. This work was co-funded by the Czech Science Foundation under Project 24-12360S, and the Grant Agency of the CTU in Prague under Project SGS24/096/OHK3/2T/13, and the European Union under the projects ROBOPROX (Robotics and Advanced Industrial Production reg. no. CZ.02.01.01/00/22\_008/0004590) and the EU Horizon Europe project XSCAVE (ID 101189836).}%
}

\begin{document}

\maketitle
\copyrightnotice


\begin{abstract}

We investigated the performance of existing semi- and fully autonomous methods for controlling flipper-based skid-steer robots. Our study involves the reimplementation of these methods for a fair comparison, and it introduces a novel semi-autonomous control policy that provides a compelling trade-off among current state-of-the-art approaches. We also propose new metrics for assessing cognitive load and traversal quality and offer a benchmarking interface for generating Quality-Load graphs from recorded data. Our results, presented in a 2D Quality-Load space, demonstrate that the new control policy effectively bridges the gap between autonomous and manual control methods. Additionally, we reveal a surprising fact that fully manual, continuous control of all six degrees of freedom remains highly effective when performed by an experienced operator on a well-designed analog controller from a third-person view.
\end{abstract}

\section{INTRODUCTION}
\input{01_intro}

\section{RELATED WORK}
\input{02_sota}


\section{METHOD}
\input{03_method}

\section{EXPERIMENTS AND RESULTS}
\input{04_experiments}

\input{06_appendix}

\section{CONCLUSIONS}
\input{05_conclusion}




\addtolength{\textheight}{-0cm}   


\bibliographystyle{plain} 
\bibliography{root}

\end{document}

%% file: 01_intro.tex
Flipper-based robots combine the strengths of both wheeled and legged systems, offering superior mobility, stability, and adaptability in challenging environments. 
Since manual control of four independent flippers during terrain
traversal is extremely demanding, even for an experienced
operator, numerous semi-/fully autonomous methods have been proposed
\cite{Kono2024, Zimmermann2014, Teymur2022, Pecka2016, Xu2024, Rocha2023}. Although these methods aim to reduce the operator's cognitive load, explicit experimental evaluations on this aspect are often lacking. The assumption that automating flipper control always reduces operator load is prevalent, yet our field experiments reveal that:

\emph{Autonomous flipper control (AFC) can be so unreliable that the added cognitive load required to manage the rest of the robot—compensating for AFC's inaccuracies—can be nearly as high as the effort needed for fully manual flipper control.} 

Furthermore, it is often erroneously assumed that fully manual real-time control of all six degrees of freedom (four flippers, heading, and speed) is technically unfeasible for a single operator. Contrary to this belief, our results demonstrate that:

\emph{Fully manual flipper control (MFC), performed by an experienced operator using a 6-axis analog controller, still achieves the best traversal performance.} 

To study this phenomenon in detail, we have reimplemented representative methods and compared their traversal quality and cognitive load on the same training data and the same test tracks. We believe that this explicit comparison provides valuable insights for the advancement of autonomous flipper control. Moreover, we propose a novel control policy that bridges the gap between existing AFC/MFC methods. This approach provides a compelling trade-off between traversal quality and operator cognitive load, delivering near-optimal traversal quality while significantly reducing the load compared to fully manual control.


As no universally accepted definition of cognitive load or traversal quality exists, we review existing metrics and introduce our definitions suited to the AFC/MFC evaluation framework. Cognitive load (CL) refers to the mental resources required to monitor, control, and navigate the robot through a given environment. CL can be measured in various ways, including subjective measures such as the NASA-Task Load Index (NASA-TLX)~\cite{hart1986nasatlx}, objective measures such as heart rate, and performance measures such as response time or error rate. In contrast to these traditional metrics, we propose measuring CL by the number of executed teleoperation commands, as it can serve as a simple but efficient proxy for the mental effort an operator expends during teleoperation. 

\begin{figure}[t!]
  \centering
  \includegraphics[width=3in]{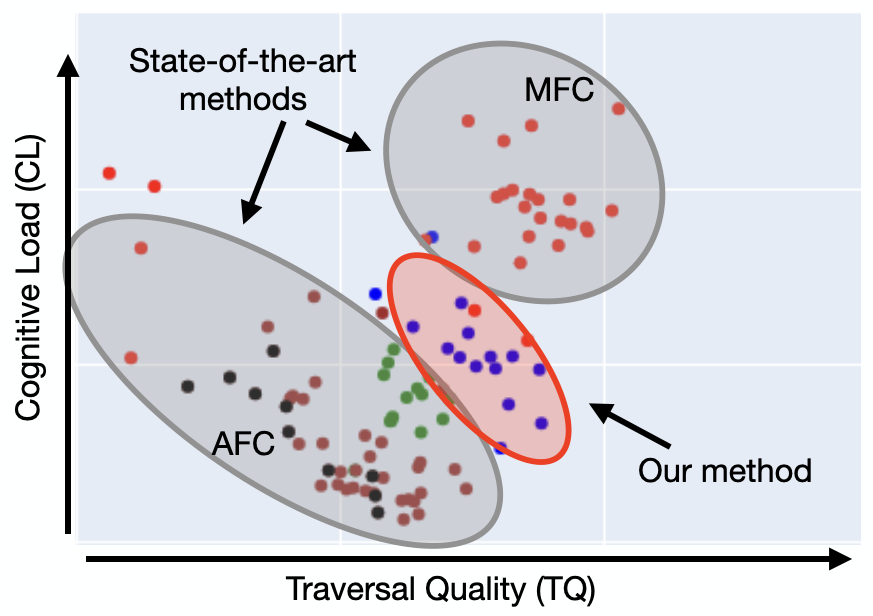}
  \caption{\textbf{Cognitive load vs traversal quality:} State-of-the-art methods outlined by grey clusters, the proposed method outlined by the red cluster. All methods manually control the speed and heading of the robot, which counts against the cognitive load. \textcolor{blue}{$\bullet$} Semi-autonomous flipper control [Ours],  \textcolor{Brown}{$\bullet$} Manual flipper control - discrete modes~\cite{Larochelle-IJRA-2013}, 
     modes~\cite{Larochelle-IJRA-2013} + anti-stuck~\cite{Teymur2022}, \textcolor{red}{$\bullet$} Manual flipper control - continuous, \textcolor{black}{$\bullet$} Autonomous flipper control - discrete modes~\cite{Larochelle-IJRA-2013} + anti-stuck~\cite{Teymur2022},
     \textcolor{Green}{$\bullet$} Autonomous flipper control - continuous~\cite{pan-2023}.}
  \label{fig:TQ-CL_graph}
\end{figure}

To quantify traversal quality (TQ), we integrate widely recognized metrics such as traversal speed and smoothness. With these cognitive load and traversal quality definitions, we present the results in a 2D Quality-Load graph (see Figure~\ref{fig:TQ-CL_graph} for the achieved outcomes).

Our analysis reveals that state-of-the-art methods form two distinct clusters on the quality-load graph: one characterized by high cognitive load and high traversal quality and the other by low cognitive load and low traversal quality. To bridge this gap, we propose a novel semi-autonomous method, which introduces prior domain knowledge in contrast to existing AFC methods.
The resulting substantial inductive bias significantly improves generalization while reducing cognitive load.



The contributions of this work are as follows:
\begin{enumerate}
\item We provide a~comprehensive comparison of various state-of-the-art methods in terms of traversal quality and cognitive load, and we identify their systematic failure cases.  
\item We introduce a novel semi-autonomous control policy that presents a compelling trade-off between existing approaches.
\item We demonstrate that the fully manual control of all 6 DOFs of the robot is cognitively manageable for an experienced operator.
\item We provide a benchmarking interface for the community that generates Quality-Load graphs based on recorded trajectories and operator commands\footnote{\url{https://github.com/ctu-vras/robot-rodeo-gym}}.

\end{enumerate}



%% file: 02_sota.tex
Flipper-based skid steer robots can be divided into two main construction categories. \emph{Six-track} robots \cite{Teymur2022, Zimmermann2014, Xu2024}, which consist of four (typically small) flippers that support the robot during uneven terrain traversal and two main tracks that control its heading and velocity. \emph{Four-track} robots \cite{Rocha2023} that consist of only four (typically heavier) flippers that handle both tasks. The latter construction is simpler, lighter, and provides slightly greater maneuverability. The former construction provides a lower risk of getting stuck on an obstacle due to its contact with the robot's body. We used a four-track construction in all our experiments because it is considered to be more challenging from the control point of view.

\emph{Manual and semi-autonomous flipper control} allows the operator to control the flippers remotely using some type of physical controller. The most straightforward way is to have fully independent control of all four flippers through an analog controller. Even though independent control of 6 degrees of freedom can seem like a big challenge, we demonstrate that such a solution still has a lot to offer compared to more sophisticated methods. To decrease cognitive load, several predefined fixed flipper positions are often used instead \cite{Zimmermann2014, Teymur2022, Larochelle-IJRA-2013}.  These predefined positions are often called \emph{traversal modes}.

\emph{Fully-autonomous flipper control} manages the flippers independently, without allowing the operator to intervene or adjust their position.
There are two major types of AFC policy architecture: (i) classifier, which classifies the robot states in discrete traversal modes, and (ii) regressor, which directly maps the robot states on continuous flipper positions. Both architectures can be trained through imitation learning~\cite{Teymur2022} or reinforcement learning~\cite{Kono2024, Zimmermann2014, Takamiya2023}. Reinforcement learning is often sample-inefficient and produces highly volatile results, whereas imitation learning delivers more consistent, deterministic results but requires an experienced operator to provide precise ground-truth terrain traversals. Given our focus on deterministic and easily repeatable results, we trained all evaluated architectures using imitation learning. 

\begin{figure}[t!]
  \centering
  \includegraphics[width=3in]{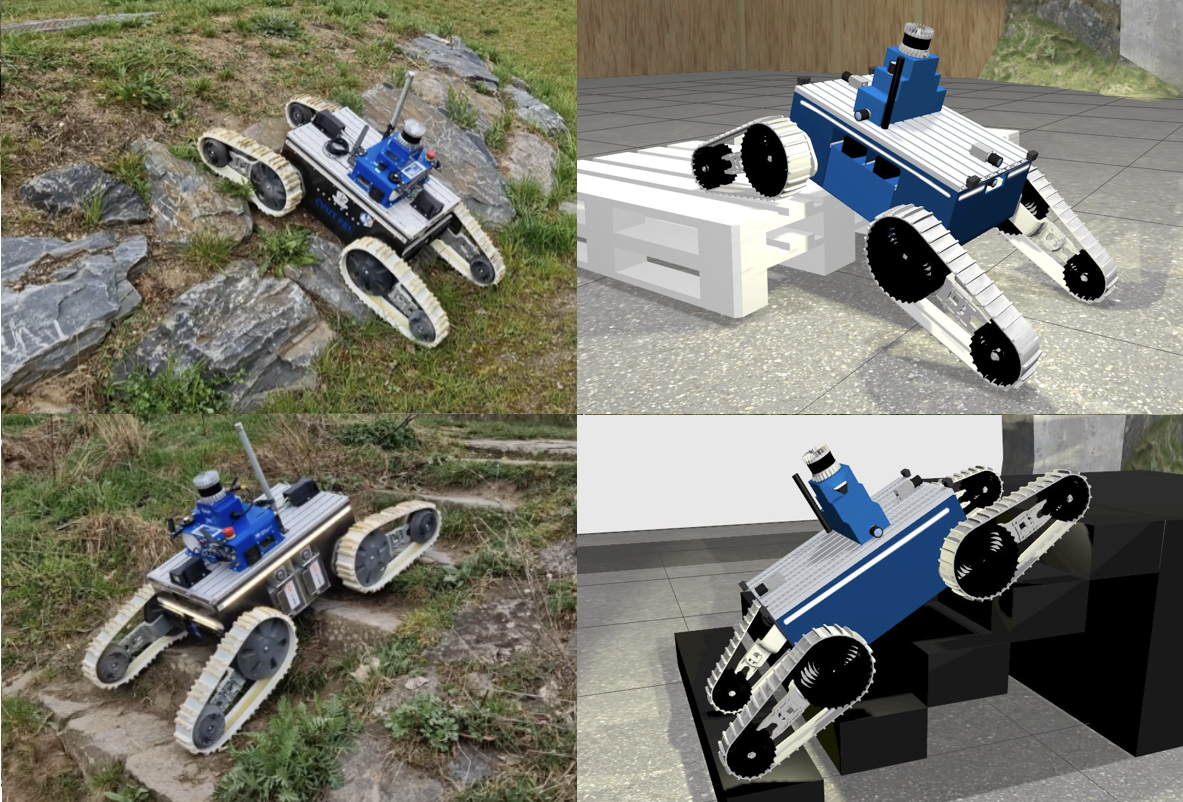}
  \caption{\textbf{Robot used for experimental evaluation:} Left shows our custom-built skid-steer robot with four independently controlled flippers. The right column displays our simulator, which employs a belt simulation instead of the multiple wheels commonly used in publicly available simulators.}
  \label{fig:robot}
\end{figure}

\emph{Training domain} is another important issue that determines the overall behavior.
Simulated and real-world environments remain popular choices for training flipper control policies. The real-world environment~\cite{Zimmermann2014, Teymur2022} reduces the gap between training and testing, but limits the number of interactions and poses risks to the robot. On the other hand, training in simulation~\cite{Kono2024, Takamiya2023} introduces a significant simulation bias, especially when exteroceptive sensors are employed.  Since AFC methods often endanger the robot during the traversal of complicated obstacles, we trained and evaluated all methods in a simulated environment. To minimize the sim2real gap, we work with an advanced simulation technique, which is based on~\cite{Pecka-Corr-2017}. The real and simulated versions of our custom-built robot are shown in Figure~\ref{fig:robot} for details.

\emph{Quantitative evaluation:} Teleoperation of robots is typically evaluated in terms of traversal quality and operator cognitive load. Hart~\cite{hart1986nasatlx} introduced the NASA Task Load Index (NASA-TLX), which assesses mental and physical demands by asking individuals to rate their experiences in six dimensions, including mental, physical, and temporal demands, as well as frustration. Larochelle et al.~\cite{Larochelle-IJRA-2013} proposed a method in which operators directly rate their cognitive load on a 1-5 scale. Fernandez et al.~\cite{FernandezRojas2020} proposed to measure the operator's heart rate to express his cognitive load; however, we argue that such a measure has too slow dynamics to capture the cognitive load connected with the traversal of a particular obstacle. In contrast to others, we propose measuring cognitive load by the number of executed teleoperation commands, as it has previously been identified to be strongly related to cognitive load~\cite{Labonte-TSMC-2010}. Many others~\cite{Xu2024, Zimmermann2014} simply assume that the operator load is decreased when the autonomous method is used and do not evaluate it at all. We argue that it is an invalid assumption since, in many cases, we observed that operators expended a lot of mental resources to compensate for failures of underlying methods.












%% file: 03_method.tex
This section is organized as follows. In~\ref{sec:method_sota}, we summarize the current state-of-the-art methods, which have been reimplemented for comparison purposes. Second, the proposed policy is described in~\ref{sec:our_policy}. Third, the normalization techniques used for the comparison are described in~\ref{sec:metrics_normalization}. Finally, we define the normalized cognitive load (\ref{sec:cognitive_load}) and normalized traversal quality~(\ref{sec:traversal_quality}).

\subsection{Summary of reimplemented state-of-the-art baselines}\label{sec:method_sota}
Various AFC/MFC methods have been proposed and implemented on different platforms, training procedures, and datasets. To ensure a fair comparison, we reimplemented a representative version of each method, using the same training procedure and dataset, and evaluated them within the same platform and environment. In some cases, we made minor adjustments to optimize performance in our testing setup. 
The robot's heading and velocity are manually controlled in all of the reimplemented flipper control methods.

\begin{figure}[!ht]
  \centering
  \includegraphics[width=\linewidth]{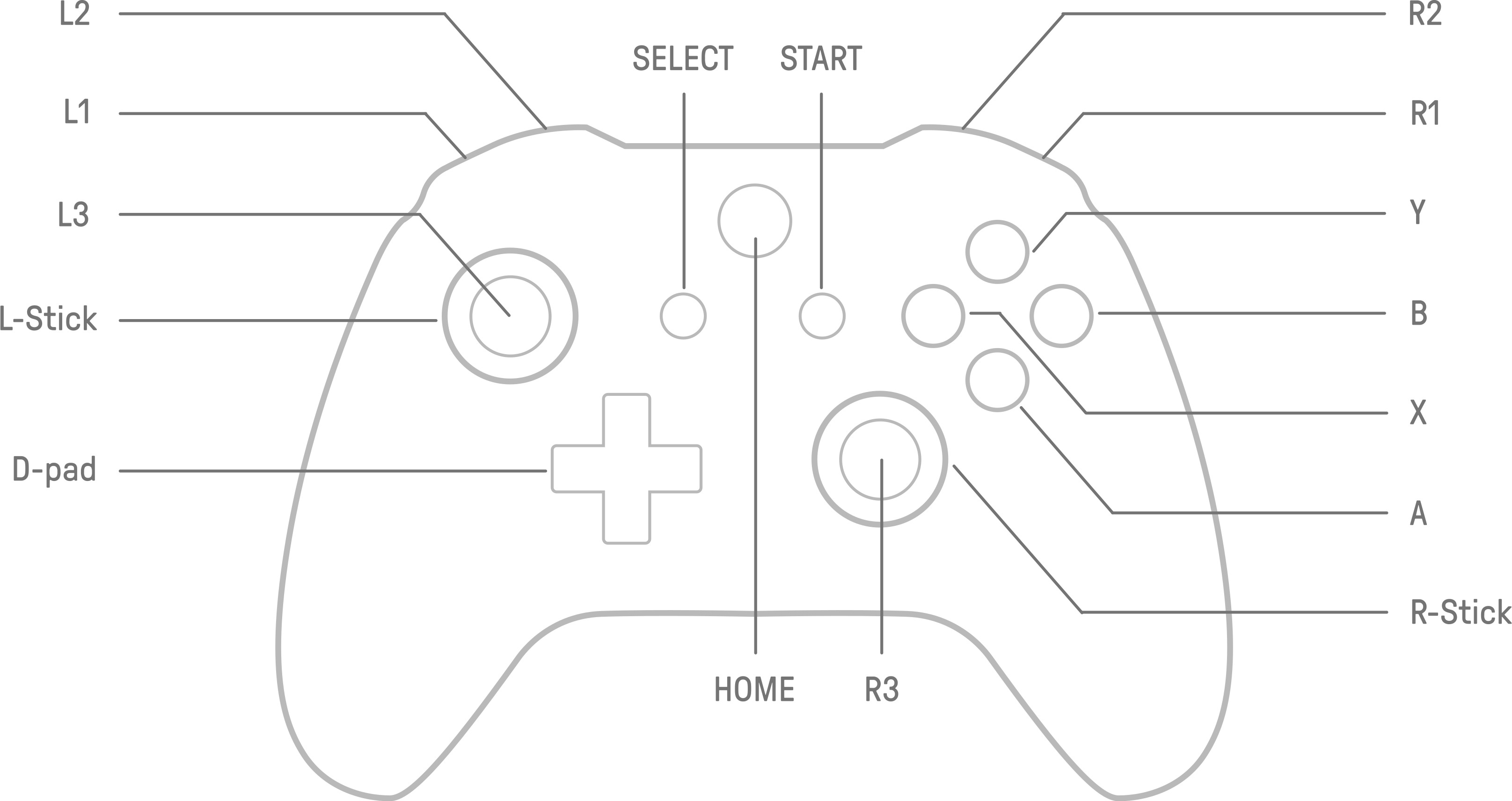}
  \caption{\textbf{Xbox Series X Controller mapping:} L-Stick controls ride movement, while R-Stick controls flipper tilting when combined with L1 (front left), R1 (front right), L2 (rear left), or R2 (rear right) buttons.}
  \label{fig:xbox_controller}
\end{figure}

\textbf{Manual flipper control - continuous:} In this method, the four flippers are manually controlled using a 6-axis remote controller; see Figure \ref{fig:xbox_controller}. To the best of our knowledge, this approach has not been previously employed, as independent real-time control of 6 continuous dimensions (4 flippers, heading, and velocity) has been considered cognitively intractable for an operator. We identified two crucial factors that enable the efficient use of this method: (i) the use of a controller with six independent analog axes and (ii) experienced operators with sufficient time for training with this controller. We observed that when inexperienced operators used this method, their results were clearly inferior to those of other existing methods -- see four red dots on the left side of the graph in Figure~\ref{fig:TQ-CL_graph}, which correspond to low-quality and high cognitive load results achieved by inexperienced operators.

\begin{figure}
    \centering
    \includegraphics[width=\linewidth]{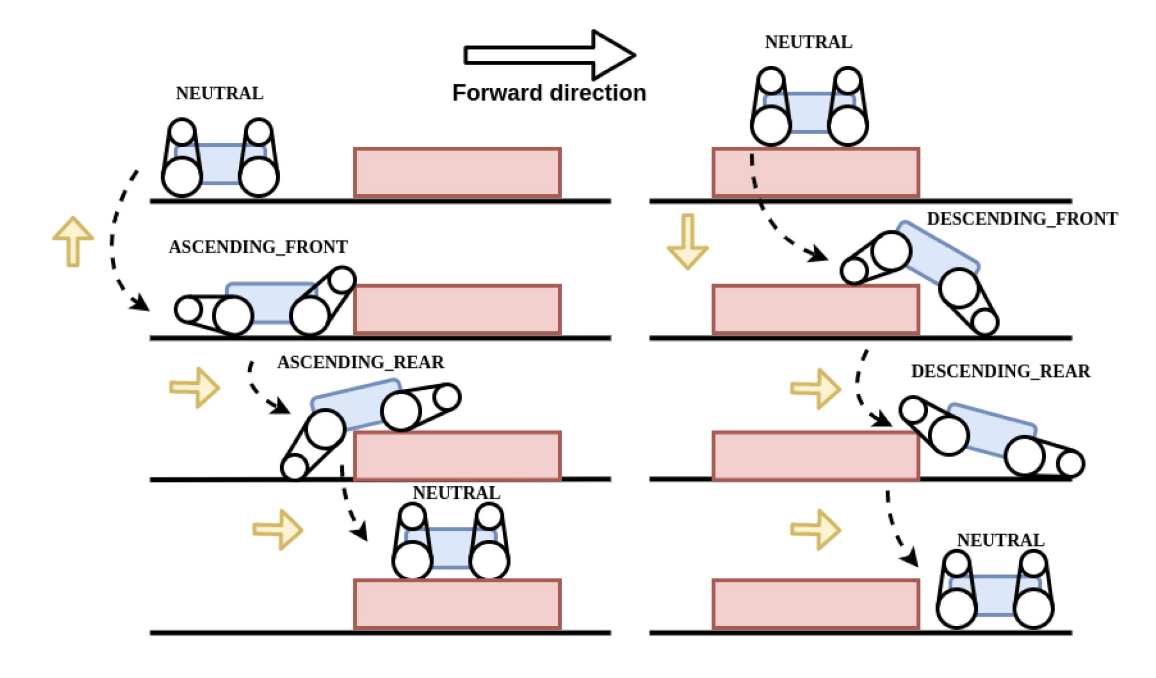}
    \caption{\textbf{Typical traversal modes:} The figure is presented with the permission of authors~\cite{Teymur2022}.}
    \label{fig:discrete_modes}
\end{figure}

\textbf{Manual flipper control - discrete modes~\cite{Larochelle-IJRA-2013}:} This method enables the operator to select from five discrete flipper configurations, which have been identified as crucial for traversing common obstacles; see Figure~\ref{fig:discrete_modes} for an example.

\textbf{Manual flipper control - discrete modes~\cite{Larochelle-IJRA-2013} + anti-stuck~\cite{Teymur2022}:} This method is an enhanced version of the previous one, introducing an additional mode that overrides the operator when the robot is detected to be stuck (i.e., ground speed is zero despite nonzero forward velocity is commanded). In this ``stuck mode'', all flippers are moved downward to lift the robot and free it from the ground.

\textbf{Autonomous flipper control - continuous~\cite{pan-2023}:} This method is an automated version of MFC-continuous policy. It employs a neural network, which controls the flipper position given the current state of the robot. Similar to~\cite{pan-2023}, our state includes the terrain heightmap transformed into the robot's coordinate frame, the flipper angles, and the robot's roll and pitch angles.  We use the same definition of state for all autonomous methods. In contrast to~\cite{pan-2023}, who train this architecture by reinforcement learning, we employ imitation learning for this architecture to ensure comparability with other methods.

\textbf{Autonomous Flipper Control - discrete modes + anti-stuck~\cite{Teymur2022}:} This method is an automated version of ``MFC-discrete modes'' policy. In contrast to the previous method, the neural network as a classifier that switches between discrete modes. This method is extended with anti-stuck mode~\cite{Teymur2022}.

\subsection{Proposed domain-aware policy}\label{sec:our_policy}
The policy
consists of two crucial components that leverage substantial prior domain knowledge. The first component, \emph{Ground-Clearance Flipper Controller (GCFC)}, controls the robot's flippers to prevent its body from getting stuck on solid ground. The second component, \emph{Obstacle-Aware Flipper Controller (OAFC)}, controls the flippers to ensure smooth entry on a frontal obstacle. During semi-autonomous teleoperation, the operator switches between these two components. In future work, we plan to train a classifier through imitation learning that will switch between these two components similarly to the presented AFC methods. The evaluated policy allows switching between the GCFC and OAFC methods for the two front flippers. Rear flippers are always controlled by GCFC.

\paragraph{Ground-Clearance Flipper Controller}
To prevent getting stuck, the algorithm maintains a minimum default distance $d_d$ between the robot's bottom and the ground. The actual minimal distance $d$ from the ground is obtained from a grid height map \cite{Fankhauser2018ProbabilisticTerrainMapping}. The desired flipper velocity $\dot{\theta}$ is then calculated using a proportional regulator:
\begin{equation}
    \dot{\theta} = p \cdot (d_d - d)\:.
\end{equation}

On flat ground, the default distance $d_d$
represents the distance between the robot's body and the ground. In all experiments, a fixed value for $d_d$ is used, with a maximum flipper angle of $\pm\pi/2$.

\paragraph{Obstacle-Aware Flipper Controller}
The front flippers are controlled to align with the most inclined normal vectors in the heightmap cells immediately in front of them. These normal vectors are obtained from a low-resolution grid height map \cite{Fankhauser2018ProbabilisticTerrainMapping}, which is intentionally chosen to filter out minor terrain disturbances and provide a smooth representation of the terrain.

\subsection{Metrics normalization}\label{sec:metrics_normalization}
To compare metrics, we normalize data by using sigmoid and linear functions.

\subsubsection{Sigmoid normalization function}\label{sec:sig_norm}
\begin{equation}
N(x, \hat{x}) = 2 - \Big(\frac{2}{1 + e^{-\frac{x}{\hat{x}}}} \Big)
    \label{eq:sigmoid_norm}
\end{equation}
%
Properties of function are when $x\rightarrow0$ then $N()\rightarrow1$ and when $x\rightarrow\infty$ then $N()\rightarrow0$. Notably, when the metric equals $\hat{x}$, the function attains a value approximately midway within the range [0, 1]. Here, $\hat{x}$ serves as a reference value against which the metric is compared.

\subsubsection{Linear normalization function}\label{sec:lin_norm}
\begin{equation}
    L(x, x_{\min}) = \max\Big(\frac{x}{x_{\min}}, 0\Big)
    \label{eq:linear_norm}
\end{equation}
Properties of function are $x\rightarrow0$ then $L()\rightarrow0$ and $x\rightarrow x_{\min}$ then $L()\rightarrow1$.

\subsection{Normalized cognitive load of the operator}\label{sec:cognitive_load}
The cognitive load of the operator is defined as the total sum over all manual interventions - button presses:
\begin{equation}
    \mathit{CL} = \sum_i \sum_j^m b_{ji} \cdot \Delta t_{i,i-1}
\end{equation}
$\Delta t_{i,i-1}$ are durations between two consecutive timestamps in time vector $T$. Let $\vec{b}_i = [b_{1i}, b_{2i}, \ldots, b_{mi}]$ be the vector representing the state of $m$ buttons and axes at time step $i$. The element $b_{ji} = 1$ when the button or axis was pressed at time $i$. Normalized cognitive load of the operator $\mathit{CL}_n$ is computed using (\ref{eq:sigmoid_norm}):
\begin{equation}
    \mathit{CL}_n = N(\mathit{CL}, \mathit{CL}_{\min})\label{eq:norm_load}
\end{equation}
where $\mathit{CL}_{\min}$ is the minimum required cognitive load for a given obstacle, which is obtained from teleoperated rides regardless of traversal quality. 

\subsection{Normalized traversal quality}\label{sec:traversal_quality}
We focus on two key metrics: robot shock and the distance of the robot body from the ground. Shock is calculated from the acceleration vector measured by IMU:
\begin{equation}
    s = \sqrt{a_x^2 + a_y^2 + a_z^2}
\end{equation}
For normalization of measured shock $s_n$, we use the sigmoid normalization (\ref{eq:sigmoid_norm}):
\begin{equation}
    s_{n} = N\Big(s, \frac{s_{\max}}{2}\Big),
\end{equation}
where $s_{\max}$ is the maximal shock obtained experimentally from the simulator. 
To normalize the distance of the robot body from the ground between [0, 1], we use normalization (\ref{eq:linear_norm})
\begin{equation}
    d_n = L(d, d_{d})
\end{equation}
This metric penalizes with $0$ when the robot bottom touches the ground ($d=0$). 
Normalized traversal quality is then the mean between these two metrics:
\begin{equation}
    \mathit{TQ}_n = \frac{s_n+d_n}{2}\label{eq:norm_quality}
\end{equation}

%% file: 04_experiments.tex
We evaluated operator load and traversal quality on 6 different flipper control methods in an arena implemented in the Gazebo physics simulator~\cite{Gazebo}. The simulator was set to have a real-time factor equal to one to reflect teleoperation in the real world. We compare our method with best-effort reimplementations of similar previous works~\cite{Larochelle-IJRA-2013}~\cite{pan-2023}~\cite{Teymur2022}~\cite{zimmermann-ICRA-2015}. To evaluate the methods in a repeatable and safe environment, we use a digital twin of our custom-built skid-steer robot with flippers; see Figure~\ref{fig:robot}. Unlike typical simulators that approximate tracks with multiple wheels, we employ a precise track simulation, leveraging conveyor belt behavior for greater accuracy. 

\begin{figure}[thpb]
  \centering
  \includegraphics[width=\linewidth]{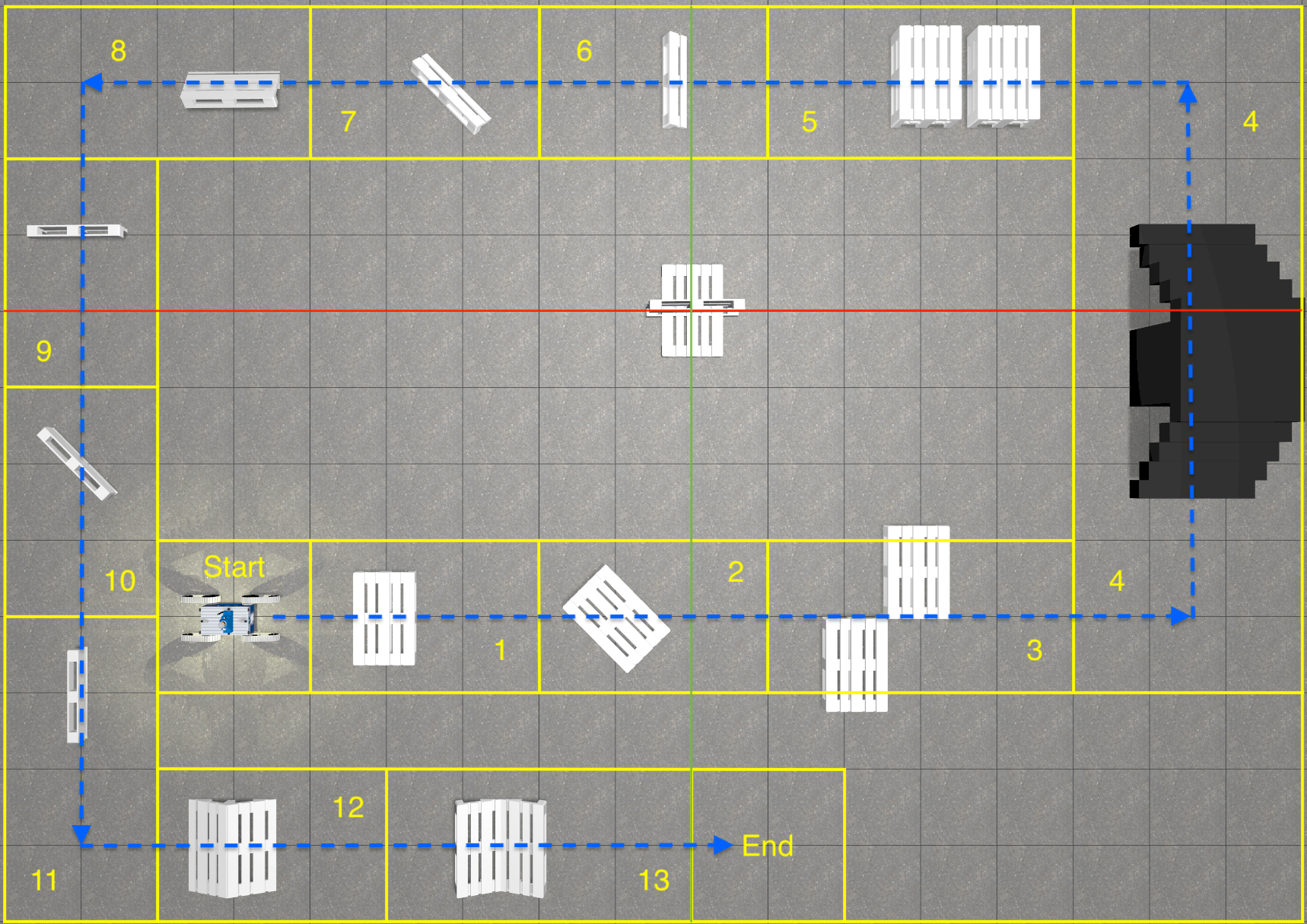}
  \caption{\textbf{Gazebo Testing arena:} arena implemented in simulator Gazebo~\cite{Gazebo} consisting of 13 various obstacles including an up-down staircase \cite{Teymur2022}.}
  \label{fig:arena}
\end{figure}

For comparison, we designed an arena with 13 challenging obstacles for traversing; see Figure~\ref{fig:arena}. Obstacles were created from simple euro pallets and up-down staircase \cite{Teymur2022}. Obstacle 3 aims to test the switch of supporting flippers in the middle of the obstacle. Three obstacles (2, 7, 9) are rotated to test how the methods can accommodate asymmetric terrain that is not parallel to the robot front. The up-down staircases are there to test the staircase traversal abilities. Obstacle 5 tests the climbing capabilities of the methods. It consists of two stacks of double pallets. Obstacles from 6 to 11 are designed to test methods on obstacles that are smaller than the robot. Tilted pallets are partially buried in the ground. The last two obstacles are ramps with A and U shapes.

To better capture operator traversal quality over the obstacle, we downsample the accelerations and distance of the robot body along the recorded traversal trajectory by dividing samples into 10 windows and selecting the maximal value within each window. Sectors are from 3 to 9 meters in length. We chose to divide samples into 10 windows from each sector. The length of the windows ranges from 0.3 to 0.9 meters. This will better capture the overall quality instead of the maximal value that is used by other methods \cite{Teymur2022, Xu2024, Rocha2023}. For example, if an operator makes a mistake at the beginning of the obstacle but then performs flawlessly for the rest of the obstacle, our metric will better reflect their overall performance rather than penalizing them disproportionately for the initial mistake. Metrics were collected by an experienced operator who made their best effort on each obstacle. Any method is penalized $\mathit{TQ}_n = 0$ and $\mathit{CL}_n = 1$ when it fails to traverse an obstacle. Metrics are categorized according to the coordinates in the arena to distinguish the data between obstacles. We then applied normalization functions (\ref{eq:norm_quality}) and (\ref{eq:norm_load}) to the collected metrics. 


Average $\mathit{CL}_n$ and $\mathit{TQ}_n$ are summarized in table \ref{table:method_comparsion}. The summary table points out that the best $\mathit{TQ}_n=0.77$ was achieved by the ``MFC - continuous'' method and followed closely by the ``Semi-AFC - continuous'' method with $\mathit{TQ}_n=0.74$. Best method with lowest $\mathit{CL}_n = 0.356$ was ``MFC - discrete modes \cite{Larochelle-IJRA-2013} + anti-stuck \cite{Teymur2022}''. As expected, methods that failed to overcome obstacles performed the worst due to penalization.

From table \ref{table:shock_comparsion} is clear that the smoothest ride was performed by the method ``AFC \cite{Teymur2022} - discrete modes \cite{Larochelle-IJRA-2013} + anti-stuck \cite{Teymur2022}''. 

The selected button mapping for the ``MFC - continuous'' method  (Fig. \ref{fig:xbox_controller}) has the drawback of not allowing different velocities to be applied to individual flippers simultaneously. The most cognitively demanding obstacles for manual control were obstacles 8 (0.76) and 10 (0.69) (see Table \ref{table:CL_comparsion}), due to their complex shapes. These obstacles require a sophisticated flipper choreography; however, the third-person view always results in one or more flippers being occluded by the robot. This is the reason why manual control does not excel in shock metrics (see first row in Table \ref{table:shock_comparsion}). It was challenging to maintain contact between all flippers and the ground. First row in Table \ref{table:distance_comparsion} shows that ``MFC - continuous'' excels in maintaining distance between the robot bottom and the ground.

In the ``MFC - discrete modes'' method, the unused buttons on the controller (Fig. \ref{fig:xbox_controller}) were used to select flipper configurations (Fig. \ref{fig:discrete_modes}). Because of very few predefined flipper positions we were not able to traverse obstacles 5, 8, 9, 10 a 11; see Table~\ref{table:method_comparsion}. On the other hand, we found that it was cognitively impossible to increase the number of modes even for an experienced operator.

Discrete modes enhanced with anti-stuck mode allows to traverse previously failed obstacles. One problem with anti-stuck mode is that it requires the robot to be stuck, which usually means a collision of the robot's bottom with the ground (Table~\ref{table:distance_comparsion}). Second problem is that once the robot is freed, control is returned to the network, which may result in the robot re-engaging the ground and potentially getting stuck again. This problem was observed at obstacles 8, 10, and 11. All methods utilizing anti-stuck suffer the same problems.


%% file: 06_appendix.tex
\begin{table*}[!ht]
\centering
\resizebox{\textwidth}{!}{
\begin{tabular}{|l|c|c|c|c|c|c|c|c|c|c|c|c|c|c|c|}
\hline
Method & Mean $\mathit{CL}_n$ $\downarrow$ & Mean $\mathit{TQ}_n$ $\uparrow$ & 1 & 2 & 3 & 4 & 5 & 6 & 7 & 8 & 9 & 10 & 11 & 12 & 13 \\ 
\hline
MFC - continuous & 0.553 & \textbf{0.774} & $\checkmark$ & $\checkmark$ & $\checkmark$ & $\checkmark$ & $\checkmark$ & $\checkmark$ & $\checkmark$ & $\checkmark$ & $\checkmark$ & $\checkmark$ & $\checkmark$ & $\checkmark$ & $\checkmark$ \\ 
\hline
MFC - discrete modes \cite{Larochelle-IJRA-2013} & 0.644 & 0.373 & $\checkmark$ & $\checkmark$ & $\checkmark$ & $\checkmark$ & $\times$ & $\checkmark$ & $\checkmark$ & $\times$ & $\times$ & $\times$ & $\times$ & $\checkmark$ & $\times$ \\ 
\hline
MFC - discrete modes \cite{Larochelle-IJRA-2013} + anti-stuck \cite{Teymur2022} & \textbf{0.356} & 0.723 & $\checkmark$ & $\checkmark$ & $\checkmark$ & $\checkmark$ & $\checkmark$ & $\checkmark$ & $\checkmark$ & $\checkmark$ & $\checkmark$ & $\checkmark$ & $\checkmark$ & $\checkmark$ & $\checkmark$ \\ 
\hline
Semi-AFC - continuous [Ours] & 0.400 & \underline{0.738} & $\checkmark$ & $\checkmark$ & $\checkmark$ & $\checkmark$ & $\checkmark$ & $\checkmark$ & $\checkmark$ & $\checkmark$ & $\checkmark$ & $\checkmark$ & $\checkmark$ & $\checkmark$ & $\checkmark$ \\ 
\hline
AFC \cite{Teymur2022} - discrete modes \cite{Larochelle-IJRA-2013} + anti-stuck \cite{Teymur2022} & \underline{0.357} & 0.704 & $\checkmark$ & $\checkmark$ & $\checkmark$ & $\checkmark$ & $\checkmark$ & $\checkmark$ & $\checkmark$ & $\checkmark$ & $\checkmark$ & $\checkmark$ & $\checkmark$ & $\checkmark$ & $\checkmark$ \\ 
\hline
AFC - continuous \cite{pan-2023} & 0.683 & 0.322 & $\checkmark$ & $\checkmark$ & $\times$ & $\times$ & $\times$ & $\checkmark$ & $\times$ & $\times$ & $\checkmark$ & $\checkmark$ & $\checkmark$ & $\times$ & $\times$ \\ 
\hline
\end{tabular}
}
\caption{\textbf{Overall methods comparison:} Average normalized cognitive load and traversal quality across all 13 obstacles and failure cases $\times$. \textbf{Bold} - best performace. \underline{Underline} - second best performace.}
\label{table:method_comparsion}

\centering
\resizebox{\textwidth}{!}{
\begin{tabular}{|l|c|c|c|c|c|c|c|c|c|c|c|c|c|c|}
\hline
Method & Mean $s_n$ $\uparrow$ & 1 & 2 & 3 & 4 & 5 & 6 & 7 & 8 & 9 & 10 & 11 & 12 & 13 \\ 
\hline
MFC - continuous & 0.68 & 0.74 & \textbf{0.78} & 0.58 & 0.58 & 0.57 & 0.73 & 0.70 & 0.65 & \underline{0.76} & 0.60 & \underline{0.76} & 0.72 & 0.71 \\ 
\hline
MFC - discrete modes \cite{Larochelle-IJRA-2013} & 0.38 & 0.80 & 0.54 & \underline{0.70} & 0.55 & $\times$ & \underline{0.80} & 0.72 & $\times$ & $\times$ & $\times$ & $\times$ & 0.77 & $\times$ \\ 
\hline
MFC - discrete modes \cite{Larochelle-IJRA-2013} + anti-stuck \cite{Teymur2022} & \underline{0.73} & \underline{0.81} & 0.61 & 0.63 & \textbf{0.70} & \underline{0.78} & \textbf{0.81} & 0.70 & \underline{0.73} & 0.68 & \underline{0.75} & 0.76 & \textbf{0.79} & \underline{0.79} \\ 
\hline
Semi-AFC - continuous & 0.73 & 0.80 & 0.67 & 0.69 & \underline{0.65} & 0.67 & 0.80 & \textbf{0.77} & 0.70 & 0.74 & \textbf{0.81} & 0.70 & \underline{0.78} & 0.73 \\ 
\hline
AFC \cite{Teymur2022} - discrete modes \cite{Larochelle-IJRA-2013} + anti-stuck \cite{Teymur2022} & \textbf{0.75} & \textbf{0.83} & \underline{0.75} & \textbf{0.70} & 0.63 & \textbf{0.79} & 0.75 & \underline{0.72} & \textbf{0.74} & \textbf{0.77} & 0.73 & \textbf{0.80} & 0.75 & \textbf{0.82} \\ 
\hline
AFC - continuous \cite{pan-2023} & 0.32 & 0.79 & 0.57 & $\times$ & $\times$ & $\times$ & 0.77 & $\times$ & $\times$ & 0.75 & 0.56 & 0.72 & $\times$ & $\times$ \\ 
\hline
\end{tabular}
}
\caption{\textbf{Normalized shock comparison:} Measured normalized shock of the robot across all 13 obstacles.  Failure cases $\times$ are penalized with lowest $s_n=0$. \textbf{Bold} - best performance. \underline{Underline} - second best performance.}
\label{table:shock_comparsion}

\centering
\resizebox{\textwidth}{!}{
\begin{tabular}{|l|c|c|c|c|c|c|c|c|c|c|c|c|c|c|}
\hline
Method & Mean $d_n$ $\uparrow$ & 1 & 2 & 3 & 4 & 5 & 6 & 7 & 8 & 9 & 10 & 11 & 12 & 13 \\ 
\hline
MFC - continuous & \textbf{0.86} & \textbf{0.90} & \textbf{0.94} & \textbf{0.90} & \underline{0.70} & \textbf{0.85} & 0.82 & \textbf{0.94} & \textbf{0.86} & \underline{0.78} & \textbf{0.77} & \textbf{0.91} & \textbf{0.89} & \textbf{0.95} \\ 
\hline
MFC - discrete modes \cite{Larochelle-IJRA-2013} & 0.37 & 0.58 & 0.62 & 0.73 & 0.45 & $\times$ & \textbf{0.87} & 0.79 & $\times$ & $\times$ & $\times$ & $\times$ & 0.77 & $\times$ \\ 
\hline
MFC - discrete modes \cite{Larochelle-IJRA-2013} + anti-stuck \cite{Teymur2022} & 0.71 & 0.75 & 0.65 & \underline{0.87} & \textbf{0.75} & 0.49 & 0.81 & 0.78 & 0.64 & 0.74 & 0.64 & 0.65 & \underline{0.87} & 0.61 \\ 
\hline
Semi-AFC - continuous & \underline{0.74} & 0.82 & \underline{0.78} & 0.64 & 0.63 & 0.69 & \underline{0.84} & \underline{0.81} & 0.66 & 0.78 & 0.69 & 0.76 & 0.83 & \underline{0.76} \\ 
\hline
AFC \cite{Teymur2022} - discrete modes \cite{Larochelle-IJRA-2013} + anti-stuck \cite{Teymur2022} & 0.66 & \underline{0.84} & 0.62 & 0.60 & 0.51 & \underline{0.75} & 0.66 & 0.51 & \underline{0.79} & 0.62 & \underline{0.72} & 0.73 & 0.53 & 0.65 \\ 
\hline
AFC - continuous \cite{pan-2023} & 0.32 & 0.70 & 0.46 & $\times$ & $\times$ & $\times$ & 0.69 & $\times$ & $\times$ & \textbf{0.82} & 0.67 & \underline{0.87} & $\times$ & $\times$ \\ 
\hline
\end{tabular}
}
\caption{\textbf{Normalized distance comparison:} Measured normalized distance of the robot bottom from the ground across all 13 obstacles. Failure cases $\times$ are penalized with lowest $d_n=0$. \textbf{Bold} - best performance. \underline{Underline} - second best performance.}
\label{table:distance_comparsion}

\centering
\resizebox{\textwidth}{!}{
\begin{tabular}{|l|c|c|c|c|c|c|c|c|c|c|c|c|c|c|}
\hline
Method & Mean $\mathit{TL}_n$ $\uparrow$ & 1 & 2 & 3 & 4 & 5 & 6 & 7 & 8 & 9 & 10 & 11 & 12 & 13 \\ 
\hline
MFC - continuous & \textbf{0.77} & \underline{0.82} & \textbf{0.86} & \underline{0.74} & 0.64 & \underline{0.71} & 0.78 & \textbf{0.82} & \underline{0.76} & \underline{0.77} & 0.69 & \textbf{0.84} & 0.81 & \textbf{0.83} \\ 
\hline
MFC - discrete modes \cite{Larochelle-IJRA-2013} & 0.37 & 0.69 & 0.58 & 0.72 & 0.50 & $\times$ & \textbf{0.84} & 0.76 & $\times$ & $\times$ & $\times$ & $\times$ & 0.77 & $\times$ \\ 
\hline
MFC - discrete modes \cite{Larochelle-IJRA-2013} + anti-stuck \cite{Teymur2022} & 0.72 & 0.78 & 0.63 & \textbf{0.75} & \textbf{0.73} & 0.64 & 0.81 & 0.74 & 0.68 & 0.71 & 0.69 & 0.70 & \textbf{0.83} & 0.70 \\ 
\hline
Semi-AFC [Ours] & \underline{0.74} & 0.81 & \underline{0.72} & 0.66 & \underline{0.64} & 0.68 & \underline{0.82} & \underline{0.79} & 0.68 & 0.76 & \textbf{0.75} & 0.73 & \underline{0.81} & \underline{0.74} \\ 
\hline
AFC \cite{Teymur2022} - discrete modes \cite{Larochelle-IJRA-2013} + anti-stuck \cite{Teymur2022} & 0.70 & \textbf{0.84} & 0.69 & 0.65 & 0.57 & \textbf{0.77} & 0.71 & 0.61 & \textbf{0.76} & 0.69 & \underline{0.73} & 0.76 & 0.64 & 0.73 \\ 
\hline
AFC - continuous \cite{pan-2023} & 0.32 & 0.74 & 0.51 & $\times$ & $\times$ & $\times$ & 0.73 & $\times$ & $\times$ & \textbf{0.79} & 0.62 & \underline{0.80} & $\times$ & $\times$ \\ 
\hline
\end{tabular}
}
\caption{\textbf{Normalized traversal quality comparison:} Measured traversal quality across all 13 obstacles. Failure cases $\times$ are penalized with lowest $\mathit{TL}=0$. \textbf{Bold} - best performance. \underline{Underline} - second best performance.}
\label{table:quality_comparsion}

\centering
\resizebox{\textwidth}{!}{
\begin{tabular}{|l|c|c|c|c|c|c|c|c|c|c|c|c|c|c|}
\hline
Method & Mean $\mathit{CL}_n$ $\downarrow$& 1 & 2 & 3 & 4 & 5 & 6 & 7 & 8 & 9 & 10 & 11 & 12 & 13 \\ 
\hline
MFC - continuous & 0.55 & 0.57 & 0.59 & 0.57 & 0.56 & 0.54 & 0.53 & 0.52 & 0.76 & 0.42 & 0.69 & 0.37 & 0.53 & 0.53 \\ 
\hline
MFC - discrete modes \cite{Larochelle-IJRA-2013} & 0.64 & 0.42 & 0.46 & 0.28 & \underline{0.35} & $\times$ & \underline{0.28} & 0.33 & $\times$ & $\times$ & $\times$ & $\times$ & \textbf{0.26} & $\times$ \\ 
\hline
MFC - discrete modes \cite{Larochelle-IJRA-2013} + anti-stuck \cite{Teymur2022} & \textbf{0.36} & 0.48 & \textbf{0.30} & \textbf{0.22} & \textbf{0.33} & \textbf{0.32} & 0.39 & 0.32 & \underline{0.64} & \textbf{0.27} & 0.29 & \underline{0.31} & 0.42 & \textbf{0.36} \\ 
\hline
Semi-AFC [Ours] & 0.40 & \textbf{0.35} & 0.51 & 0.32 & 0.44 & \underline{0.37} & 0.36 & \underline{0.30} & 0.68 & 0.35 & 0.34 & 0.32 & 0.40 & 0.45 \\ 
\hline
AFC \cite{Teymur2022} - discrete modes \cite{Larochelle-IJRA-2013} + anti-stuck \cite{Teymur2022} & \underline{0.36} & \underline{0.35} & \underline{0.32} & \underline{0.26} & 0.46 & 0.42 & 0.39 & \textbf{0.19} & \textbf{0.54} & 0.35 & \textbf{0.20} & 0.44 & \underline{0.28} & \underline{0.43} \\ 
\hline
AFC - continuous \cite{pan-2023} & 0.68 & 0.43 & 0.41 & $\times$ & $\times$ & $\times$ & \textbf{0.27} & $\times$ & $\times$ & \underline{0.28} & \underline{0.23} & \textbf{0.26} & $\times$ & $\times$ \\ 
\hline
\end{tabular}
}
\caption{\textbf{Normalized cognitive load comparison:} Measured cognitive load for each method across all 13 obstacles. Failure cases $\times$ are penalized with highest $\mathit{CL}_n=1$. \textbf{Bold} - best performance. \underline{Underline} - second best performance.}
\label{table:CL_comparsion}

\end{table*}

%% file: 05_conclusion.tex
We implemented representative methods of Manual Flipper Control (MFC) and Autonomous Flipper Control (AFC) and compared them in terms of traversal quality and operator cognitive load. We also introduced a novel policy that utilizes straightforward yet effective domain knowledge and provides a compelling trade-off between quality and cognitive load. Specifically, this policy achieves over $96\%$ of the best traversal quality while reducing the cognitive load by approximately $30\%$. 

We also found that the best traversal quality ($0.77$) is still achieved by an experienced operator with continuous real-time control of all 6 degrees of freedom of the robot at the cost of a high cognitive load ($0.55$). The best fully autonomous method is~\cite{Teymur2022}. It achieves a similar traversal quality ($0.72$) but with nearly half the cognitive load ($0.36$). We also identified that their anti-stuck system may suffer from oscillations. In particular, the anti-stuck system successfully frees the robot and returns the control to the policy network, which often pushes the body back on the obstacle and potentially gets stuck again. This issue recurs, particularly with obstacles such as long, narrow logs or walls. On the other hand, our experiments show that augmenting an arbitrary method with the anti-stuck mode is sufficient to traverse all obstacles, although with low traversal quality.

Furthermore, our analysis shows that poorly performing fully autonomous policies not only exhibit inferior traversal quality but also result in significantly higher cognitive load. This is because the unreliability of autonomous flipper control requires increased cognitive effort to manage the robot's heading and velocity, outweighing any cognitive load savings achieved through automation.






%% file: root.bbl
\begin{thebibliography}{10}

\bibitem{Teymur2022}
Teymur Azayev and Karel Zimmermann.
\newblock Autonomous state-based flipper control for articulated tracked robots in urban environments.
\newblock {\em IEEE Robotics and Automation Letters}, 7(3):7794--7801, 2022.

\bibitem{Fankhauser2018ProbabilisticTerrainMapping}
P{\'{e}}ter Fankhauser, Michael Bloesch, and Marco Hutter.
\newblock Probabilistic terrain mapping for mobile robots with uncertain localization.
\newblock {\em IEEE Robotics and Automation Letters (RA-L)}, 3(4):3019--3026, 2018.

\bibitem{FernandezRojas2020}
Raul Fernandez~Rojas, Eloise Debie, Jack Fidock, Michael Barlow, Kathryn Kasmarik, Sreenatha Anavatti, Matthew Garratt, and Hussein Abbass.
\newblock Electroencephalographic workload indicators during teleoperation of an unmanned aerial vehicle shepherding a swarm of unmanned ground vehicles in contested environments.
\newblock {\em Frontiers in Neuroscience}, 14:40, 2020.

\bibitem{hart1986nasatlx}
Sandra~G. Hart.
\newblock {NASA Task Load Index (TLX): Paper and Pencil Package}, January 1 1986.

\bibitem{Gazebo}
Nathan Koenig and Andrew Howard.
\newblock Design and use paradigms for gazebo, an open-source multi-robot simulator.
\newblock In {\em IEEE/RSJ International Conference on Intelligent Robots and Systems}, pages 2149--2154, Sendai, Japan, Sep 2004.

\bibitem{Kono2024}
Hitoshi Kono, Sadaharu Isayama, Fukuro Koshiji, Kaori Watanabe, and Hidekazu Suzuki.
\newblock Automatic flipper control for crawler type rescue robot using reinforcement learning.
\newblock {\em International Journal of Advanced Computer Science and Applications}, 15(6), 2024.

\bibitem{Labonte-TSMC-2010}
Daniel Labonte, Patrick Boissy, and Fran\c{c}ois Michaud.
\newblock Comparative analysis of 3-d robot teleoperation interfaces with novice users.
\newblock {\em Trans. Sys. Man Cyber. Part B}, 40(5):1331–1342, oct 2010.

\bibitem{Larochelle-IJRA-2013}
Benoit Larochelle, Geert-Jan Kruijff, and Jurriaan van Diggelen.
\newblock Usage of autonomy features in usar human-robot teams.
\newblock {\em International Journal of Intelligent Systems and Applications in Robotics (IJRA)}, 4(1):19--30, June 2013.

\bibitem{pan-2023}
Hainan Pan, Xieyuanli Chen, Junkai Ren, Bailiang Chen, Kaihong Huang, Hui Zhang, and Huimin Lu.
\newblock Deep reinforcement learning for flipper control of tracked robots in urban rescuing environments.
\newblock {\em Remote Sensing}, 15(18), 2023.

\bibitem{Pecka-Corr-2017}
Martin Pecka, Karel Zimmermann, and Tomáš Svoboda.
\newblock Fast simulation of vehicles with non-deformable tracks.
\newblock In {\em 2017 IEEE/RSJ International Conference on Intelligent Robots and Systems (IROS)}, pages 6414--6419, 2017.

\bibitem{Pecka2016}
Martin Pecka, Vojtěch Šalanský, Karel Zimmermann, and Tomáš Svoboda.
\newblock Autonomous flipper control with safety constraints.
\newblock In {\em 2016 IEEE/RSJ International Conference on Intelligent Robots and Systems (IROS)}, pages 2889--2894, 2016.

\bibitem{Rocha2023}
Filipe Rocha, Andre Cid, Mário Delunardo, Renato~P. Junior, Nilton~S Thiago, Luiz Barros, Jacó~D. Domingues, Gustavo Pessin, Gustavo Freitas, and Ramon Costa.
\newblock Body posture controller for actively articulated tracked vehicles moving over rough and unknown terrains.
\newblock In {\em 2023 IEEE/RSJ International Conference on Intelligent Robots and Systems (IROS)}, pages 2330--2337, 2023.

\bibitem{Takamiya2023}
Hidenori Takamiya, Ryosuke Yajima, Jun Younes~Louhi Kasahara, Ren Komatsu, Keiji Nagatani, Atsushi Yamashita, and Hajime Asama.
\newblock Reinforcement learning-based motion generation for a tracked robot to go over a sphere-shaped non-fixed obstacle.
\newblock In {\em 2023 IEEE/SICE International Symposium on System Integration (SII)}, pages 1--6, 2023.

\bibitem{Xu2024}
Zhengzhe Xu, Yanbo Chen, Zhuozhu Jian, Junbo Tan, Xueqian Wang, and Bin~Liang Liang.
\newblock Hybrid trajectory optimization for autonomous terrain traversal of articulated tracked robots.
\newblock {\em IEEE Robotics and Automation Letters}, 9(1):755--762, 2024.

\bibitem{Zimmermann2014}
Karel Zimmermann, Petr Zuzanek, Michal Reinstein, and Vaclav Hlavac.
\newblock Adaptive traversability of unknown complex terrain with obstacles for mobile robots.
\newblock In {\em 2014 IEEE International Conference on Robotics and Automation (ICRA)}, pages 5177--5182, 2014.

\bibitem{zimmermann-ICRA-2015}
Karel Zimmermann, Petr Zuzánek, Michal Reinstein, Tomáš Petříček, and Václav Hlaváč.
\newblock Adaptive traversability of partially occluded obstacles.
\newblock In {\em 2015 IEEE International Conference on Robotics and Automation (ICRA)}, pages 3959--3964, 2015.

\end{thebibliography}
